\documentclass[11pt,a4paper]{article}
\usepackage[hyperref]{emnlp-ijcnlp-2019}
\usepackage{times}
\usepackage{latexsym}
\usepackage{mlsymbols}
\usepackage{mystyle}
\usepackage{symbol}
\usepackage{comment}

\usepackage{url}

\aclfinalcopy 

\title{A Simple Global Neural Discourse Parser}

\author{
    Yichu Zhou  \\
    University of Utah \\
    {\tt flyaway@cs.utah.edu} \\\And%
    Omri Koshorek \\
    Tel-Aviv University \\
    {\tt omri.koshorek@cs.tau.ac.il} \\\AND
    Vivek Srikumar \\
    University of Utah \\
    {\tt svivek@cs.utah.edu} \\\And%
    Jonathan Berant \\
    Tel-Aviv University\\
    {\tt joberant@cs.tau.ac.il}
  }

\date{}

\begin{document}
\maketitle
\begin{abstract}
    Discourse parsing is largely dominated by
    greedy parsers with manually-designed
    features, while global parsing is rare due to its
    computational expense.  In this paper, we propose a
    simple chart-based neural discourse parser that does not
    require any manually-crafted features and is based on
    learned span representations only. To overcome the
    computational challenge, we propose an independence
    assumption between the label assigned to a node in the
    tree and the splitting point that separates its children,
    which results in tractable decoding. We empirically
    demonstrate that our model achieves the best performance
    among global parsers, and comparable performance to
    state-of-art greedy parsers, using only learned
    span representations.
\end{abstract}

\section{Introduction}\label{sec:intro}
The discourse structure of a document describes discourse
relationships between its elements as a graph or a tree. Discourse
parsing is largely dominated by greedy
parsers~\cite[\eg.][]{braud2016multi,ji2014representation,yu2018transition,SogaardBC17}.
Global parsing is rarer~\cite{joty2015codra,li2016discourse} because
the dependency between node's label and its internal split point can
make prediction computationally prohibitive.

In this work, we propose a CKY-based global parser with tractable inference using a new
independence assumption that loosens the coupling between the identification of
the best split point label prediction.
Doing so gives us the advantage that we can search
for the best tree in a larger space.
Greedy discourse
parsers~\cite{braud2016multi,ji2014representation,yu2018transition,SogaardBC17}
have to use complex models to ensure each step is
correct because the search space is limited.
For example, \citet{ji2014representation} 
manually crafted features and feature transformations to
encode elementary discourse units (EDUs); \citet{yu2018transition} and
\citet{braud2016multi} used multi-task learning for
a better EDU representation.
Instead, in this work, we use a simple recurrent span
representation to build a parser that outperforms previous  global
parsers.

Our contributions are:
\begin{inparaenum}[(i)]
\item We propose an independence assumption that allows global
  inference for discourse parsing.
\item Without any manually engineered features, our {\em simple}
  global parser outperforms previous global methods for the task.
\item Experiments reveal that our parser outperforms greedy
  approaches that use the same representations, and is comparable to
  greedy models that rely on hand-crafted features or more data.
\end{inparaenum}


\section{RST Tree Structure}\label{sec:rst}
\begin{figure}
    \centering
    \includegraphics[width=0.4\textwidth]{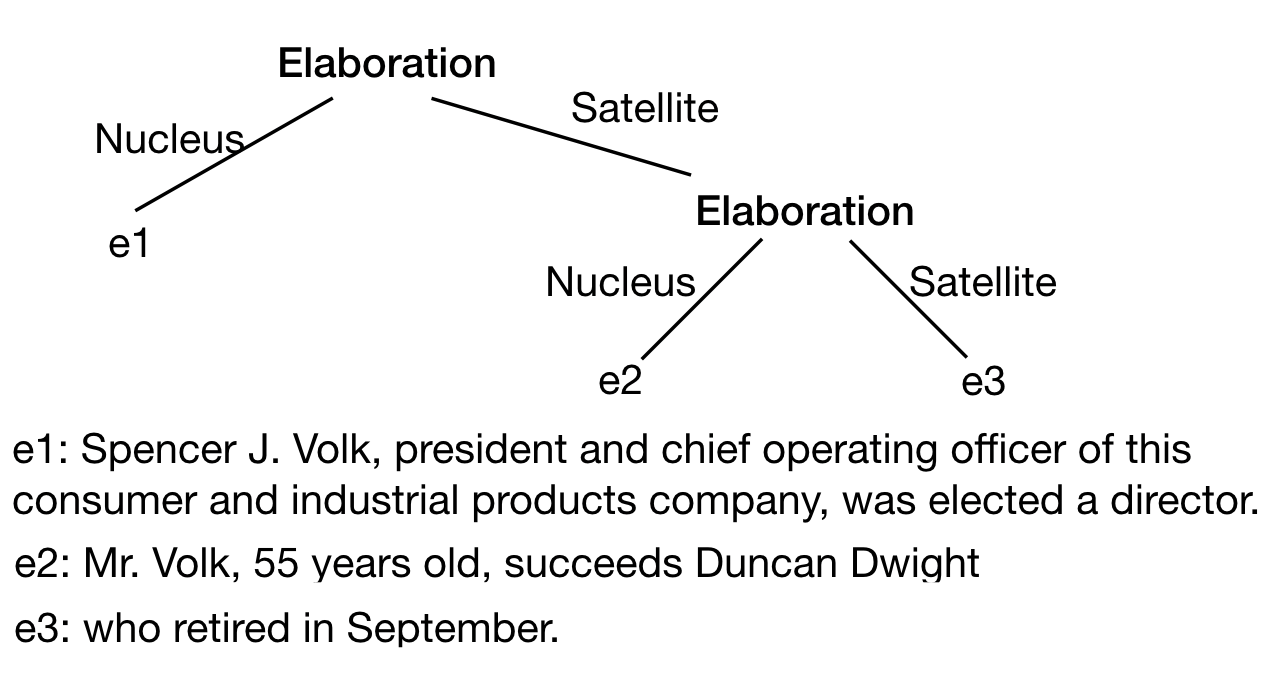}
    \caption{An example of RST tree, where $\{e_1,e_2,e_3\}$
    are EDUs, \textit{Elaboration} is a discourse relation
    label. \textit{Nucleus} and \textit{Satellite} are
    nuclearity labels.}\label{fig:example} \end{figure}

The Rhetorical Structure theory (RST) of~\citet{mann1988rhetorical}
is an influential theory on discourse.
In this work, we focus on discourse parsing with the RST
Discourse Treebank~\cite{carlson2001building}. 
An RST tree
assigns relation and nuclearity labels
to adjacent nodes.  Leaves, called  elementary discourse units (EDUs),
are clauses (not words) that serve as building blocks
for RST trees. Figure~\ref{fig:example} shows an example RST
tree.

RST trees have important structural differences from constituency
parse trees.
In a constituency tree, node labels describe their syntactic
role in a sentence, and are independent of the splitting
point between their children, thus driving methods such as
that of~\citet{DBLP:conf/acl/SternAK17}.  However, in an RST
tree, the label of a node describes the relationship between
its sub-trees;
the assignment of labels depends on the split point
that separates its children.

\section{Chart-based Parsing}\label{sec:chart}

In this section, we will first describe chart
parsing, and then look at our independence assumption that reduces
inference time.
Finally, we will look at the the loss function for training the parser.

\subsection{Chart Parsing System}
An RST tree structure $T$ can be represented as a set of labeled spans:
\begin{equation}
    T:= \pc{\p{\p{i,j},l,p}}
\end{equation}
where, for a span $(i,j)$, the relation label is  $l$ and the
nuclearity, which determines the direction of the relation, is $p$.
The score of a tree $S_{tree}(T)$ is the sum of its span,
relation and nuclearity scores.


To find the best tree, we can use a chart to store the scores of
possible spans and labels.  For each cell $(i,j)$ in the table, we
need to decide the splitting point $k$, and the nuclearity and
relation labels.

As we saw in \S\ref{sec:rst}, the label and split decisions are
not independent (unlike, \eg~\citet{DBLP:conf/acl/SternAK17}).  The joint score
for a cell is the sum of all three scores, and also the scores of
its best subtrees:

%
\begin{equation}
    \begin{aligned}
        S_{best}(i,j) = & \max_{k,l,p}\left[S_{span}(i,k) + S_{span}(k,j)\right. \\
                        & + S_{rel}(i,j,k,l) + S_{nuc}(i,j,k,p) \\
                        & \left.+ S_{best}(i,k) + S_{best}(k,j)\right]
    \end{aligned}
    \label{eq:exact-recursion}
\end{equation}
The base case for a leaf node does not account for
the split point and subtrees:
\begin{equation}
    \small
    S_{best}(i,i+1) = \max_{l,p}\left[S_{rel}(i,i+1,i,l) + S_{nuc}(i,i+1,i,p)\right]
\end{equation}

The CKY algorithm can be used for decoding the recursive definition
in \eqref{eq:exact-recursion}.  The running time is $O(Gn^3)$, where
$n$ is the number of EDUs in a document, and $G$ is the grammar
constant, which depends on the number of labels.

\subsection{Independence Assumption}\label{sec:independent}

Although we have framed the parsing process as a chart parsing
system, the large grammar constant $G$ makes inference expensive.
To resolve this, we assume that we can identify the splitting point
of a node without knowing the its label.
After this decision, we use this
split point to inform the label predictors instead of
searching for the best split point jointly.
The scoring function becomes:
\begin{equation}
    \begin{aligned}
        S_{best}(i,j) = & S_{span}(i,k) + S_{span}(k,j)\\
                        & +\max\limits_{l}S_{rel}(i,j,k,l) \\
                        & + \max\limits_{p}S_{nuc}(i,j,k,p) \\
                        & + S_{best}(i,k) + S_{best}(k,j)\\
        \text{where } k = &\arg\max\limits_{k}(S_{span}(i,k) + S_{span}(k,j) \\
            &+ S_{best}(i,k) + S_{best}(k,j)) 
    \end{aligned}
    \label{eq:semi-recursion}
\end{equation}

Unlike the parser of~\citet{li2016discourse} that completely disentangles
label and splitting points, we retain a one-sided dependency. The
joint score is still used in the recursion.  Because they
are not completely independent, we call our assumption the
\textbf{partial independence assumption}. When we use the CKY
algorithm as the inference algorithm to
resolve equation~\ref{eq:semi-recursion}, the running time
complexity becomes $O(n^2(n+G))$. While we still have a cubic
dependency on the number of EDUs, the impact of the
constant makes our approach practically feasible.

\subsection{Loss Function}%
\label{sec:chart-loss}
Since inference is feasible, we can train the model with inference
in the inner step. Specifically, we use a max-margin loss that is
the neural analogue of a structured
SVM~\cite{taskar2005learning}. Recall that if we had all our scoring
functions, we can predict the best tree using CKY as
\begin{equation}
  \hat{T} = \arg\max_{T}[S_{tree}(T)]
\end{equation}
For training, we can use the gold tree $T^*$ of a document to define
the structured loss as:
\begin{equation}
    \small
    \ell(T^{*}, \hat{T}) =
    \left[S_{tree}(\hat{T})+\Delta(\hat{T},T^{*})-S_{tree}(T^{*})\right]
\end{equation}
$\Delta(\hat{T},T^{*})$ is the hamming distance between a tree
$\hat{T}$ and the reference $T^{*}$.  The above loss can be computed
with loss-augmented decoding as standard for a structured SVM, thus
giving us a sub-differentiable function in the model parameters.


\section{Neural Model for Global Parser}\label{sec:rep}


In this section, we describe our neural model that defines the
scoring functions using a EDU representation. The network
first maps a document---a sequence of words $w_1, \ldots, w_n$---to
a vector representation for each EDU in the
document. Those EDU representations serve as inputs to the three
predictors: $S_{span}, S_{rel}$ and $S_{nuc}$.
 
Since the \textit{relation} and \textit{nuclearity} of a
\textit{span} depend on its context, recurrent neural networks are a
natural way of modeling the sequence, as they have been shown
successfully capture word/span context for many NLP
applications~\cite{DBLP:conf/acl/SternAK17,DBLP:journals/corr/BahdanauCB14}.

Each word $w_i$ is embedded by the concatenation of its
GloVe~\cite{pennington2014glove} and ELMo
embeddings~\cite{peters2018deep}, and embeddings of its POS
tag. These serve as inputs to a bi-LSTM network.  The POS tag
embeddings are initialized uniformly from $(0,1)$ and updated during
the training process, while the other two embeddings are not
updated.  The softmax-normalized weights and scale parameters of
ELMo are fine-tuned during the training process.

Suppose for a word  $w_i$, the forward and backward
encodings from the Bi-LSTM are $\mathbf{f}_i$ and $\bb_i$ respectively.  The
representation of an EDU with span $(i,j)$, denoted as $\be$, is the
concatenation of its encoded first and last words:
\begin{equation}
    \be = \mathbf{f}_i\oplus\bb_i\oplus\mathbf{f}_j\oplus\bb_j.
\end{equation}
The parameters of this EDU representation include three parts:
\begin{inparaenum}[(i)]
    \item POS tag embeddings;
    \item Softmax-normalized weights and scalar parameter for
        ELMo;
    \item Weights of the bi-LSTM\@.
\end{inparaenum}

Using this representation, our scoring functions \ie,
$S_{span}, S_{rel}$ and $S_{nuc}$, are implemented as a two-layer
feedforward neural network which takes an EDUs representation to
score their respective decisions. The EDU representation parameters
and the scoring functions are jointly learned.


\section{Experiments}\label{sec:exp}
The primary goal of our experiments is to compare the partial
independence assumption against the full independence assumption
of~\citet{li2016discourse}. In addition, we also compare the global
models against a shift-reduce parser (as
in~\citet{ji2014representation}) that uses the same
representation.

We evaluate our parsers on the RST Discourse
Treebank~\cite{carlson2001building}.
It consists of $385$ documents in total, with $347$ training and
$38$ testing examples. We further created a development set by
choosing $47$ random documents from the training set for development
and to fine tune hyperparameters. The supplementary material lists
all the hyperparameters.

Following previous studies~\cite{carlson2001building}, the original $78$ relation types are partitioned into $19$ classes.
All experiments are conducted on manually segmented EDUs.
The POS tag of each word in the EDUs is obtained from
spaCy\footnote{\url{https://spacy.io/}}.
We train our parser on the training split and use the
best-performing model on the
development set as the final model. We optimized the max-margin loss using Adam~\cite{kingma2014adam}. 



We use the standard evaluation method~\cite{marcu2000theory} to
test model performances using three metrics: Span,
Nuclearity and Relation (Full).
We follow \citet{morey2017much} to report both macro-averaged
and micro-averaged F1 scores.

\subsection{Results}\label{sec:results}

Table~\ref{tb:macro} shows the final performance of our
parsers using macro-averaged F1 scores. Our
partial independence assumption outperforms the complete
independence assumption by a large margin.
Among all other parsers, our partial independence parser achieves
the best results.  Table~\ref{tb:micro} shows the performance of our
parsers using micro-averaged F1 scores. Under this metric,
the partial independence assumption still outperforms the complete
independence assumption and the baseline. Again, we are among the
the best-performing parsers, though the best method \citet{yu2018transition} is
shift-reduce based parser augmented by multi-task learning. The
latter's better performance, as per in the ablation study of the
original work, is due to the use of external resources (Bi-Affine
Parser) for a better representation.

\begin{figure}[htp]
    \includegraphics[width=0.5\textwidth]{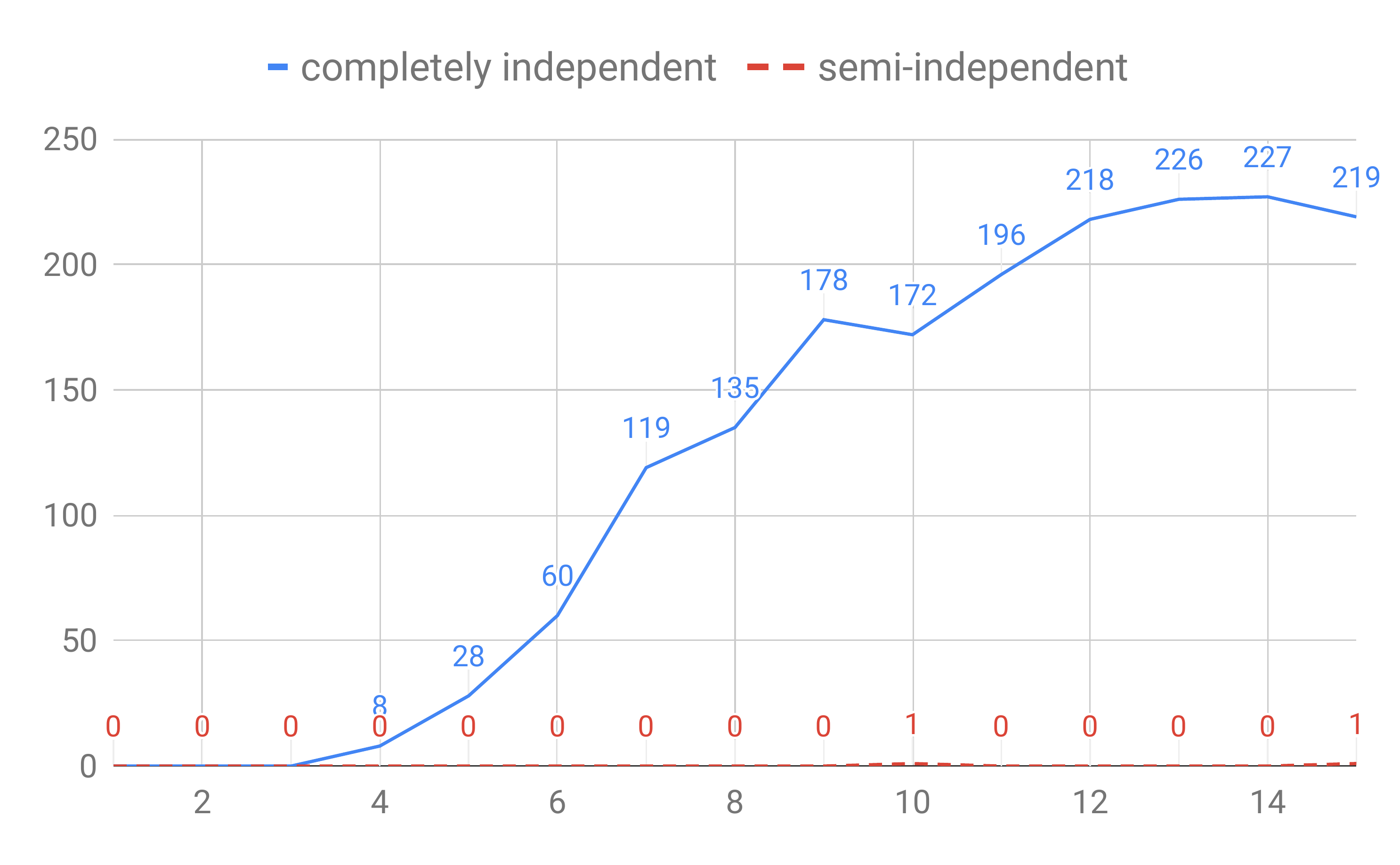}
    \caption{The number of missing prediction trees for
    different independence assumptions. X-axis is the training
    epochs, y-axis is the number of missing trees for each
    epoch.}\label{fig:negative}
\end{figure}

To better understand the difference between
complete independence and partial independence assumption,
we count how many trees that found by the inference algorithm has a
lower score than the corresponding gold tree during training. Since both assumptions cannot perform exact
search, it is possible to find a tree whose score is higher than the
gold one.
We call this situation \textbf{missing prediction}.
Figure~\ref{fig:negative} shows the results.
Complete independence assumption produces more missing
prediction trees. This is because, in complete independence
assumption, the tree structure is decided only by its span
scores. A tree can have high span scores but lower label
scores, resulting in a low score in total.

\begin{table}[htb]
\centering
\small
\begin{tabular}{@{}clccc@{}}
\toprule
    Categories & Parsing System & S & N & R \\ \midrule
    \multirow{2}{*}{Global} & \citet{joty2015codra} & 85.7 & 73.0 & 60.2 \\
    & \citet{li2016discourse} & 85.4 & 70.8 & 57.6 \\\midrule
    \multirow{6}{*}{Greedy} & \citet{SogaardBC17} & 85.1 & 73.1 & 61.4 \\
    & \citet{feng2014linear} & 87.0 & 74.1 & 60.5 \\
    & \citet{surdeanu2015two} & 85.1 & 71.1 & 59.1 \\
    & \citet{hayashi2016empirical} & 85.9 & 72.1 & 59.4 \\
    & \citet{braud2016multi} & 83.6 & 69.8 & 55.1 \\
    & \citet{ji2014representation} & 85.0 & 71.6 & 61.9 \\
    & Baseline & 86.6 & 73.8 & 61.6 \\ \midrule
    \multirow{2}{*}{Our System} & Complete Independence & 85.7 & 72.2  & 56.7 \\
    & Partial Independence & \textbf{87.2} & \textbf{74.9}  & \textbf{61.9} \\ \midrule
    & Human & 89.6 & 78.3 & 66.7 \\ \bottomrule
\end{tabular}
\caption{Macro-averaged $F_1$ comparison for different
    parsers. The results of other models are from
    \citet{morey2017much}. Baseline is a shift-reduce parser
    that uses the same representation as our system.}\label{tb:macro}
\end{table}

\begin{table}[htb]
\centering
\small
\begin{tabular}{@{}clccc@{}}
\toprule
    Categories & Parsing System & S & N & R \\ \midrule
    \multirow{2}{*}{Global}   & \citet{joty2015codra} & 82.6 & 68.3 & 55.4 \\
    & \citet{li2016discourse} & 82.2 & 66.5 & 50.6 \\ \midrule
    \multirow{8}{*}{Greedy}    & \citet{SogaardBC17} & 81.3 & 68.1 & 56.0 \\
    & \citet{feng2014linear} & 84.3 & 69.4 & 56.2 \\
    & \citet{surdeanu2015two} & 82.6 & 67.1 & 54.9 \\
    & \citet{hayashi2016empirical} & 82.6 & 66.6 & 54.3 \\
    & \citet{braud2016multi} & 79.7 & 63.6 & 47.5 \\
    & \citet{ji2014representation} & 82.0 & 68.2 & 57.6 \\
    & \citet{yu2018transition} & \textbf{85.5} &
    \textbf{73.1} & \textbf{59.9} \\
    & Baseline & 83.3 & 70.4 & 56.7 \\ \midrule
    \multirow{2}{*}{Our System}& Complete Independence & 83.0 & 67.7  & 51.8\\ 
    & Partial Independence  & 84.5 & 71.1  & 57.5 \\\midrule
    & Human & 88.3 & 77.3 & 65.4 \\ \bottomrule
\end{tabular}
\caption{Micro-averaged $F_1$ comparison for different
    parsers. The results of other models are from
    \citet{morey2017much}. Baseline is a shift-reduce parser
    that uses the same representation as our system.}\label{tb:micro}
\end{table}

\section{Analysis and Related Work}\label{sec:related}

Some prior work explores global parsing for RST structures
\citet{li2016discourse} used the CKY algorithm to infer by 
ignoring the dependency relation between
splitting point and label assignment. 
\citet{joty2015codra} applied a two-stage parsing
strategy. A sentence is first parsed, and then the document
is parsed. In this process, all the cross-sentence spans
are ignored.

Greedy parsing can only explore a small part of the output space,
thus necessitating high-quality representation and models to ensure
each step is as correct as possible.  This is the reason why many
early studies usually involve rich manually engineered
features~\cite{joty2015codra,feng2014linear}, external
resources~\cite{yu2018transition,braud2016multi} or heavily designed
models~\cite{li2016discourse,ji2014representation}.
Table~\ref{tb:components} summarize all the different components
used by various parsers.  In contrast, using global
inference, our parser only needs a recurrent input representation
to achieve comparable performance without any components mentioned
in Table~\ref{tb:components}.

\begin{table}[htb]
\centering
\scriptsize
\begin{tabular}{ccccc}
    \toprule
    \begin{tabular}[c]{@{}c@{}}Parsing \\ System\end{tabular} &
    \begin{tabular}[c]{@{}c@{}}Manual \\ Features \end{tabular} &
    \begin{tabular}[c]{@{}c@{}}Two \\ Stages \end{tabular} & Multi-task &
    \begin{tabular}[c]{@{}c@{}}Feature\\ Transform\end{tabular} \\\midrule
    \citet{joty2015codra} & $\surd$ & $\surd$ &   &  \\
    \citet{li2016discourse} & $\surd$ &  &   &  $\surd$ \\
    \citet{SogaardBC17} & $\surd$ &   &  &  \\
    \citet{feng2014linear} & $\surd$ & $\surd$ &  &  \\
    \citet{surdeanu2015two} & $\surd$ &  &   &  \\
    \citet{hayashi2016empirical} & $\surd$ &  &  &  \\
    \citet{braud2016multi} &  &  & $\surd$ &   \\
    \citet{ji2014representation} & $\surd$ &  &  & $\surd$ \\
    \citet{yu2018transition} &  &  & $\surd$ & \\
\bottomrule
\end{tabular}
\caption{Components in different parsing models in the literature. By manual 
    features, we mean human designed features other than POS
    tags. In comparison, our global parser uses none of these components.}\label{tb:components}
\end{table}


\section{Conclusion}\label{sec:conclusion}
In this work, we propose a new independence assumption for global
inference of discourse parsing, which makes globally optimal
inference feasible for RST trees.  By using a global inference, we
develop a simple neural discourse parser. Our experiments 
show that the simple parser can achieve comparable performance to
state-of-art parsers using only learned span representations.

\section*{Acknowledgements}
\label{sec:acknowledgements}

This research was supported by The U.S-Israel Binational
Science Foundation grant 2016257, its associated NSF grant
1737230 and The Yandex Initiative for Machine Learning.

\bibliography{cited}
\bibliographystyle{acl_natbib}

\newpage
\appendix
\section{Hyper-parameters for Experiments}
Table~\ref{tb:hyper} shows the hyper-parameters for 
our experiments.

\begin{table}[h]
    \centering
\begin{tabular}{lc}
\toprule
Hyper-parameters & Setting  \\ \midrule
Max Epoch &  $15$ \\
biLSTM Hidden Size  &  $200$ \\
Feedforward Hidden Size  &  $200$ \\
GloVe Word Embedding Size & $300$ \\
ELMo Word Embedding Size & $1024$ \\
POS Tag Embedding Size & $300$ \\
Dropout Probability  &  $0.2$ \\
Learning Rate  &  $0.001$ \\
\bottomrule
\end{tabular}
    \caption{Hyper-parameters in all
    experiments.}\label{tb:hyper}
\end{table}

\end{document}